%% file: main.tex
\documentclass{article}
\pdfoutput=1

\usepackage[preprint]{neurips_2024}

\usepackage[utf8]{inputenc} % allow utf-8 input
\usepackage[T1]{fontenc}    % use 8-bit T1 fonts
\usepackage{hyperref}       % hyperlinks
\usepackage{url}            % simple URL typesetting
\usepackage{booktabs}       % professional-quality tables
\usepackage{amsfonts}       % blackboard math symbols
\usepackage{nicefrac}       % compact symbols for 1/2, etc.
\usepackage{microtype}      % microtypography
\usepackage{xcolor}         % colors
\usepackage{amsmath} % for better formatting
\usepackage{enumitem} % to control list parameters
\usepackage{graphicx}
\usepackage{float}
\usepackage{multirow}
\usepackage[normalem]{ulem}
\usepackage{tabularx}
\usepackage{arydshln}

\title{Expediting and Elevating Large Language Model Reasoning via Hidden Chain-of-Thought Decoding}

\author{%
    Tianqiao Liu \\ TAL Education Group \\ Beijing, China \\  \texttt{liutianqiao1@tal.com}  \\
    \And
    Zui Chen \\ TAL Education Group \\ Beijing, China \\  \texttt{chenzui3@tal.com}  \\
    \And
  Zitao Liu \\ Jinan University \\ Guangzhou, China \\  \texttt{liuzitao@jnu.edu.cn}  \\
  \And
  Mi Tian \\  TAL Education Group \\ Beijing, China \\  \texttt{tianmi@tal.com}  \\
  \And
  Weiqi Luo \\ Jinan University \\ Guangzhou, China \\ \texttt{lwq@jnu.edu.cn} \\
}

\begin{document}

\maketitle

\begin{abstract}
\input{abstract}
\end{abstract}

\input{intro}

\input{background}

\input{method}

\input{experiment}

\input{related}
\vspace{-0.2cm}
\input{conclusion}

\bibliographystyle{acl_natbib}
\bibliography{custom}

\newpage
\onecolumn
\appendix
\input{appendix.tex}

\end{document}

%% file: abstract.tex
Large language models (LLMs) have demonstrated remarkable capabilities in tasks requiring reasoning and multi-step problem-solving through the use of chain-of-thought (CoT) prompting. However, generating the full CoT process results in significantly longer output sequences, leading to increased computational costs and latency during inference. To address this challenge, we propose a novel approach to compress the CoT process through semantic alignment, enabling more efficient decoding while preserving the benefits of CoT reasoning. Our method introduces an auxiliary CoT model that learns to generate and compress the full thought process into a compact special token representation semantically aligned with the original CoT output. This compressed representation is then integrated into the input of the Hidden Chain-of-Thought (HCoT) model. The training process follows a two-stage procedure: First, the CoT model is optimized to generate the compressed token representations aligned with the ground-truth CoT outputs using a contrastive loss. Subsequently, with the CoT model parameters frozen, the HCoT model is fine-tuned to generate accurate subsequent predictions conditioned on the prefix instruction and the compressed CoT representations from the CoT model. Extensive experiments across three challenging domains - mathematical reasoning, agent invocation, and question answering - demonstrate that our semantic compression approach achieves competitive or improved performance compared to the full CoT baseline, while providing significant speedups of at least 1.5x in decoding time. Moreover, incorporating contrastive learning objectives further enhances the quality of the compressed representations, leading to better CoT prompting and improved task accuracy. Our work paves the way for more efficient exploitation of multi-step reasoning capabilities in LLMs across a wide range of applications.

%% file: intro.tex
\section{Introduction}
Chain-of-Thought (CoT) prompting, as introduced by \citep{weiChainofThoughtPromptingElicits}, involves prompting large language models (LLMs) to generate explicit reasoning steps, significantly enhancing their performance in various reasoning tasks such as mathematical problem solving (\citealp{hendrycksMeasuringMathematicalProblem2021}, \citealp{cobbeTrainingVerifiersSolve2021}) and science question answering \citep{lu2022learn}. Subsequent CoT variants (\citealp{zhouLeasttoMostPromptingEnables2023}, \citealp{chenProgramThoughtsPrompting2022}, \citealp{gaoPALProgramaidedLanguage2023}) have aimed at further improving its efficacy across diverse domains. However, these methods enhance LLMs' reasoning capabilities by extending the duration and complexity of reasoning processes and incorporating external computational resources to achieve superior outcomes. This increased computational demand may limit their applicability in real-world development scenarios.

\begin{figure}[H]
  \centering
  \includegraphics[width=\linewidth]{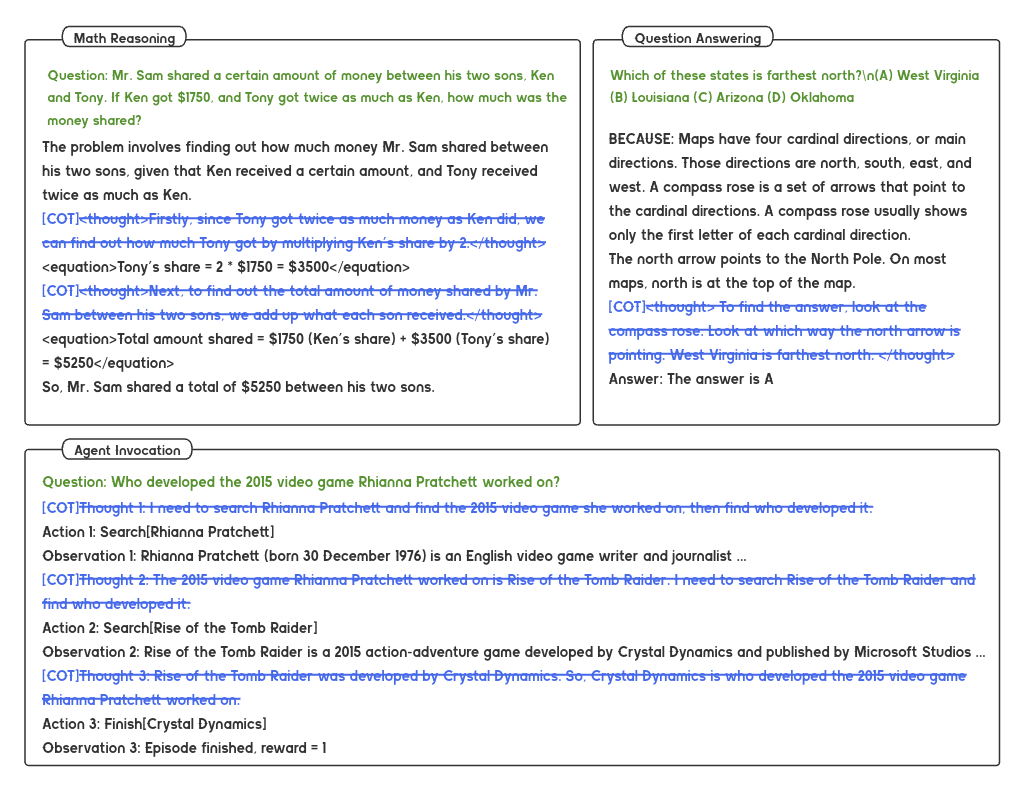}
  \vspace{-0.8cm}
  \caption{Real-world examples of the CoT prompting in tasks such as mathematical reasoning, question answering, and agent invocation. In the figure, green parts represent actual user queries, blue strikethroughs indicate compressed thought processes, and black text denotes non-CoT content, which corresponds to the expected output for users.} % 
    \label{fig:infer}
\end{figure}

As illustrated in Figure \ref{fig:infer}, the blue tokens represent intermediate reasoning steps. These steps are crucial for ensuring final decoding accuracy but contribute to significant computational overhead when using standard CoT prompting. Generating complete CoT sequences typically requires substantially longer output sequences. \cite{jinImpactReasoningStep2024} indicates that extending the reasoning steps in prompts can enhance the reasoning abilities of LLMs, even without introducing new information. Nevertheless, within the transformer architecture \citep{vaswaniAttentionAllYou2017}, decoding time increases linearly with the output length, leading to higher computational costs and increased latency during inference. These issues, while critical, have not been well addressed in the existing literature.

% \textit{``The best thoughts are those that remain unspoken'' --- Charles Dickens}
To address this challenge, we draw upon the human cognitive process, where chains of thought are often implicitly and instantaneously formed within the mind. This introspection led us to hypothesize that during the decoding phase, a single token could represent the forthcoming cognitive process, effectively compressing the semantic content of an extensive reasoning chain into a specialized token. 
This approach aligns with recent findings in the domain of In-context Learning (ICL) for LLMs. Research by \cite{wangLabelWordsAre2023} has demonstrated the feasibility of employing 'anchor tokens' as potent conduits for aggregating and transmitting complex information. 

Leveraging this foundation, we propose a novel two-stage fine-tuning framework aimed at generating subsequent outputs, such as precise answers or computational formulas, by utilizing \textbf{a compressed special token} representation in conjunction with the preceding context. The first stage of this framework involves the training of an auxiliary CoT model. This model employs a contrastive loss function to effectively condense an elaborate thought process into a specialized token, herein referred to as [CoT]. Subsequently, we fine-tune our \textbf{Hidden CoT (HCoT) model} to generate the desired output based on the representation of the special token encoded by the CoT model and the preceding instructions, with the parameters of the CoT model remaining frozen. During inference, as shown in Figure \ref{fig:infer}, our HCoT model halts upon encountering the [CoT] token, at which point it feeds the preceding information to the auxiliary CoT compression model. The auxiliary model then generates a compressed CoT representation encapsulating the subsequent thought process. This compressed representation is then reinserted into the HCoT model to complete the inference. Concurrently, the auxiliary CoT compression model can either continue to generate the full thought process or opt not to, in order to conserve computational resources. Building on the inherent parallelizability of the LLM encoding process, the encoding phase that yields the special token representation is markedly more time-efficient when compared to the time-consuming process of decoding a complete chain of thought. Consequently, this optimization significantly accelerates the rate of inference.

Our extensive experiments show the potential of HCoT method on four datasets in three challenging domains: mathematical reasoning (\citealp{hendrycksMeasuringMathematicalProblem2021}, \citealp{cobbeTrainingVerifiersSolve2021}), agent invocation \citep{yaoReActSynergizingReasoning2023}, and science question answering \citep{lu2022learn}. The results demonstrate that our HCoT model achieves competitive or improved performance compared to the full CoT baseline, while providing significant speedups of over 1.5x to 3.8x in decoding time.

To summarize, our major contributions are:
\begin{itemize}[leftmargin=10pt]
    \item We propose the Hidden Chain-of-Thought (HCoT) framework, a novel approach that accelerates the inference process of large language models by compressing the multi-step reasoning process into a specialized token representation, thereby reducing computational overhead during decoding.
    \item We introduce a disentangled training paradigm for the multi-step CoT reasoning, enabling isolated error correction and specialized optimization for each component. 
    \item Our compression model effectively condenses the entire thought process into a compact special token while maintaining interpretability, allowing for parallel generation of CoT content.
    \item By incorporating a contrastive learning objective, we further enhance the quality of the compressed CoT model. This approach improves CoT prompting and task accuracy through the application of a span-level loss function during supervised fine-tuning.
\end{itemize}
We believe our work paves the way for more efficient exploitation of multi-step reasoning capabilities in LLMs across a wide range of applications.

%% file: background.tex
\section{Background}
We first formalize some existing methods in this section. Our approach is not only inspired by these prior techniques but is also benchmarked against them for comparative analysis. We denote a pre-trained LLM with parameters $\theta$. We use lowercase letters such as $x$, $c$, and $z$ to represent the user's question, the generated content, and the CoT reasoning process, respectively. For instance, a user question is denoted as $x = (x[1], \cdots, x[n])$, where each $x[i]$ is an individual token, and the probability of the sequence under our model is given by $p_{\theta}(x) = \prod_{i=1}^{n} p_{\theta}(x[i]|x[1] ... x[i-1])$. To accommodate the complexity of reasoning that involves multiple interleaved sequences of content and thought, we extend our notation. For the \textit{i}-th element in the reasoning process, we use subscripts: $z_i$ represents the \textit{i}-th chain of thought, and $c_i$ represents the \textit{i}-th output content sequence. We use uppercase letters $C$ and $Z$ to denote a collection of output contents and thoughts respectively.

{\bf Chain-of-Thought (CoT) Reasoning} introduces intermediate steps that guide the model towards generating a more structured and potentially more accurate response. In this framework, the output is divided into several components: intermediate steps $z_i$ and content parts $c_i$. The process involves iteratively generating these components based on previous steps: where model sample \textit{i}-th thought $z_i \sim p_{\theta}(z_i \mid x, c_0, z_0 \cdots c_i)$ and then sample following content $c_{i+1} \sim p_{\theta}(c_{i+1} \mid x, c_0, z_0 \cdots c_{i}, z_{i})$ given the sampled \textit{i}-th thought.

{\bf Reasoning w/o Chain-of-Thought (CoT)} contrasts with the CoT methodology by directly generating the final answer without explicitly modeling the intermediate reasoning steps. In this approach, the model aims to produce the output content $C$ directly from the user's question $x$: $C \sim p_{\theta}(C \mid x)$. Here, the the reasoning process is implicit within the model's parameters $\theta$ and is not visible.

%% file: method.tex
\section{HCoT: Hidden Chain-of-Thought Reasoning}
\begin{figure}[]
  \centering
  \includegraphics[width=\linewidth]{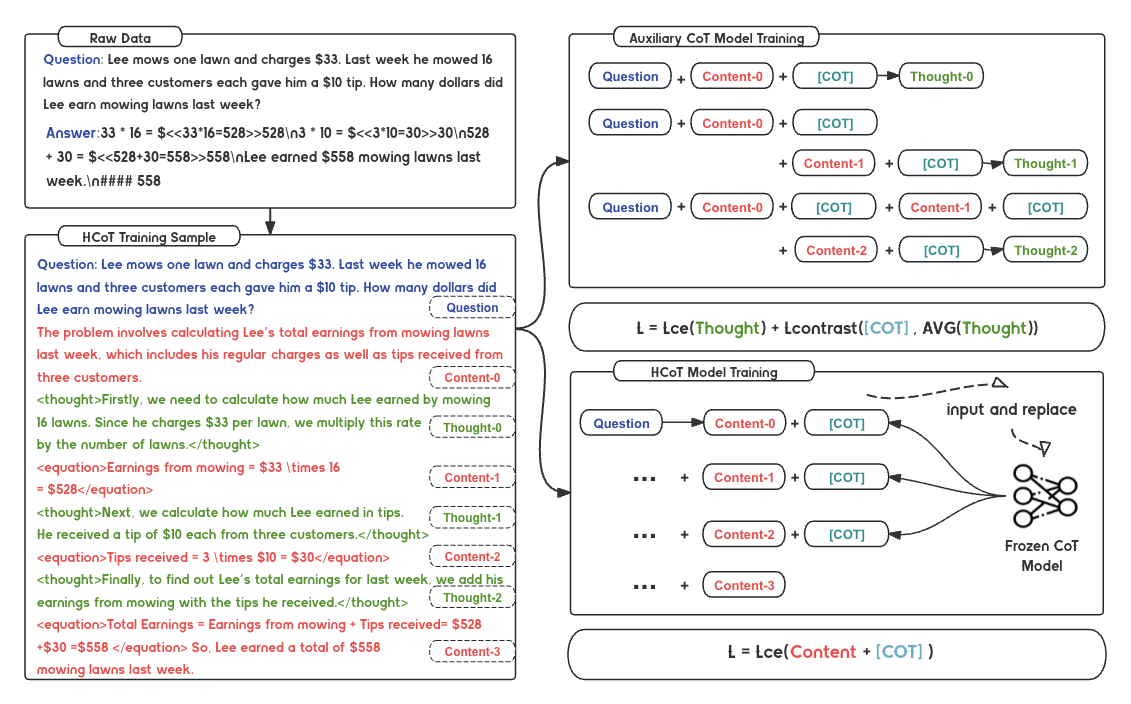}
  \vspace{-0.3cm}
  \caption{Data construction and two-stage training of Hidden Chain-of-Thought (HCoT) models for math reasoning tasks: Training instances are synthetically generated from raw data using GPT-4, then utilized separately for training the Auxiliary Chain-of-Thought Model and the HCoT Model.} % 
  %数学领域HCoT 模型的Two-stage training的 数据构建和训练过程。解释一下raw data, HCoT Training samples, CoT Model Training, LLM Model Training 之间的关系。HCoT Training samples 由Raw data使用GPT4 with instructions制作。然后它被分别用作制作CoT Model 和LLM Model的训练数据。
    \label{fig:train}
\end{figure}

In this section, we present our novel two-stage training method that incrementally develops the auxiliary CoT model followed by the Hidden CoT (HCoT) model. The auxiliary CoT model is ingeniously crafted to distill the reasoning process into a singular \texttt{[CoT]} token. Subsequently, the HCoT model leverages the encapsulated reasoning within the \texttt{[CoT]} token to facilitate swift and efficient chain-of-thought reasoning. As depicted in Figure~\ref{fig:train}, our training paradigm encompasses three components: (i) the generation of HCoT training instances from the original dataset to construct data that embodies CoT reasoning, (ii) the auxiliary CoT model training that employs these HCoT instances to create specialized training samples aimed at enhancing CoT reasoning, and (iii) the HCoT Model training phase that utilizes the same HCoT training instances, first replacing the intermediate reasoning steps $z_i$ with the special \texttt{[CoT]} token, and upon encountering this special token, leveraging the encoded hidden representation from the frozen Auxiliary CoT model to replace the original input embedding for the special \texttt{[CoT]} token in the HCoT Model, thereby guiding the generation of the subsequent content $c_{i+1}$. In the following subsections, we delineate the methodology for constructing HCoT training samples in Section~\ref{HCoT_trianing_sample}, elucidate the training dataset preparation and the training procedure for the auxiliary CoT model in Section~\ref{cot_model_train}, and detail the dataset construction and training process for the HCoT model in Section~\ref{hcot_model_train}.

\subsection{HCoT Training Sample Construction}
\label{HCoT_trianing_sample}
Constructing training samples for HCoT is a flexible process that can be tailored to the specific requirements of the task at hand and the anticipated format of the CoT. For instance, in our experiments on math reasoning tasks, as depicted in Figure 2, data from sources such as Math and GSM8K are utilized to construct training samples employing a GPT-4 based ICL method. (Specific prompts are in Appendix\ref{appen:prompts}.) This approach enables the model to output a series of contents and thoughts denoted as  $c_0, z_0, \ldots, z_{n-1}, c_n$ which is sampled from:
\begin{equation}
\label{equ:whole-cot}
p(C, Z \mid x) = \prod_{i=1}^{n} p_{\theta}(c_0 \mid x)p_{\theta}(z_{i-1} \mid x, \ldots, c_{i-1})p_{\theta}(c_i \mid x, c_0, \ldots, z_{i-1}), 
\end{equation}
This recursive formulation encapsulates the dependencies between the \textbf{intermediate reasoning steps} ($z_{i-1}$) and the content ($c_i$) that is produced as a result. The flexibility of the HCoT sample construction process allows for the adaptation of the training regime to accommodate diverse reasoning patterns and content structures. By iteratively generating and conditioning on the chain of thought, the model is trained to better align its reasoning process with the underlying logic of the task.

\subsection{Disentangled Training Paradigm}
Delving deeper into Equation~ \ref{equ:whole-cot}, we have disentangled the probability distribution into distinct components, we can identify the segment $p_{\theta}(z_{i-1} \mid x, \ldots, c_{i-1})$ as the generation phase of CoT information, namely \textbf{auxiliary CoT model}. Conversely, the segment $p_{\theta}(c_i \mid x, c_0, \ldots, z_{i-1})$ is responsible for harnessing the generated CoT to produce subsequent content, namely content generation model. This observation has led to the conceptualization of disentangled training paradigm which first compressing the high-dimensional discrete distribution of $p_{\theta}(z_{i-1} \mid x, \ldots, c_{i-1})$ into a more compact thought representation via encoding the input content with $p_{\theta}^{COT}$, which corresponding to the auxiliary CoT model trianing. Subsequently, the content generation model $p_{\theta}^{HCoT}$ is fine-tuned to maximize $p(c_i \mid x, c_0, \ldots, z_{i-1})$, where $[z_0 \ldots z_{i-1}]$ denote the compressed representations provided by the CoT model. This disentangled training paradigm offers several beneficial training dynamics:

\textbf{Error Isolation:} By decoupling the training of the auxiliary CoT model ($p_{\theta}^{COT}$) from the content generation model ($p_{\theta}^{HCoT}$), errors in reasoning can be isolated within the auxiliary CoT model. This facilitates targeted corrections without affecting the content generation model, thereby preventing the propagation of errors and enhancing the overall robustness of the system.

\textbf{Specialized Optimization:} The disentangled training paradigm allows for specialized optimization strategies. The CoT model can be honed to refine reasoning abilities and logical coherence, while the content generation model can concentrate on articulating clear and pertinent content. This specialization ensures that each model maximizes its performance in its respective domain, leading to a more effective training process.

\textbf{Parallel Development and Improved Interpretability:} The CoT generation operates in parallel with the generation of actual content. This not only accelerates the inference speed but also preserves the interpretability of the model. Unlike a black box approach, the reasoning steps generated by the CoT model are explicit and can be scrutinized, allowing for a better understanding of the model's thought process and facilitating easier debugging and refinement.

To be noticed, We also include the part of $p_{\theta}(c_0 \mid x)$ in our content generation model.

\subsection{Auxiliary CoT Model}
\label{cot_model_train}
In this section, We use lowercase letter $r$ to represent special \texttt{[CoT]} token's representation, and $r_i$ denotes the $i$-th special \texttt{[CoT]} representation. Given the user's question and the preceding content, the objective of the auxiliary CoT model is to distill the reasoning process into a compact representation by maximizing $p_{\theta}^{COT}(z_i \mid x, \ldots, c_i, r_i)$, where $z_i$ is the desired thought process.

\textbf{Training Data Configuration:} As depicted in the top right corner of Figure~\ref{fig:train}, the training data for the auxiliary CoT model is constructed by first extracting all the thought processes from the original HCoT training samples. Subsequently, between each content segment $c_i$ and each thought segment $z_i$, we insert a special token, denoted as \texttt{[CoT]}, where this unique token serves as an anchor to facilitate the generation of the subsequent thought process. We then segment the entire training sample into individual instances, each treating a thought process $z_i$ as the target output, conditioned on the preceding context comprising the question $x$, special \texttt{[CoT]} tokens and the content segments up to $c_{i-1}$.

\textbf{Thought Compression:} The auxiliary CoT model $p_{\theta}^{COT}$ is trained by maximizing the likelihood $p_{\theta}^{COT}(z_i \mid x, \ldots, c_i, r_i)$, where we expect the model to generate the most accurate thought representation based on the preceding information. In addition to the conventional cross-entropy loss, we incorporate symmetric contrastive loss between \textbf{the thought process respresentations mean-pooling} and the [CoT] token representation to enhance the thought compression capability of the model. The underlying assumption is that the compressed thought representation should exhibit a higher affinity with its corresponding special \texttt{[CoT]} token than with other \texttt{[CoT]} tokens, and vice versa. The final loss function for the auxiliary CoT model is as follows:

\begin{equation}
\begin{aligned}
\mathcal{L}_{\text{CoT}} &= \mathcal{L}_{\text{CE}} + \lambda \cdot \mathcal{L}_{\text{contrastive}} = -\log p_{\theta}^{CoT}(z_i \mid x, \ldots, c_i, r_i) \\ & - \frac{\lambda}{2} \cdot (log \frac{exp(\mathbf{z}_i \cdot \mathbf{r}_i)}{\sum_{k=0}^{n} exp(\mathbf{z}_i \cdot \mathbf{r}_k)} + log \frac{exp(\mathbf{z}_i \cdot \mathbf{r}_i)}{\sum_{j=0}^{n} exp(\mathbf{z}_j \cdot \mathbf{r}_i)}) \\
\end{aligned}
\end{equation}

, where $n$ denotes the batch size during the auxiliary CoT model's training, $\mathbf{z}_i \in \mathcal{R}^{d}$ represents the normalized representation of the $i$-th target thought process, obtained by mean-pooling the final hidden states from $p{\theta}^{COT}$ when the input is the sequence $[z_i[0], z_i[1], \ldots, z_i[t]]$. Additionally, $\mathbf{r}_i \in \mathcal{R}^{d}$ denotes the normalized representation of the corresponding special \texttt{[CoT]} token generated by the auxiliary CoT model. The $\mathcal{L}_{\text{contrastive}}$ is introduced to enhance the compactness of the thought process representation, ensuring that it exhibits a higher affinity towards the corresponding target thought process representation. Here, $\lambda$ is a hyperparameter that governs the trade-off between the contrastive loss and the primary cross-entropy loss term.

\subsection{HCoT Model}
\label{hcot_model_train}
\textbf{Training Data Configuration:} The training data construction process for the HCoT model is illustrated in the bottom right corner of Figure~\ref{fig:train}. We replace all the thought processes $z_i$ in the original HCoT training samples with the special \texttt{[CoT]} token. This transformation aligns the training paradigm seamlessly with the auxiliary CoT model's training, as they share the same input format. By substituting the explicit thought processes with the compact \texttt{[CoT]} token, we effectively leverage the distilled reasoning encapsulated within the auxiliary CoT model's output representation. This approach ensures that the HCoT model's training is closely tied to the learned thought representations from the auxiliary CoT model, facilitating the transfer of reasoning capabilities.

\textbf{Supervised Fine-tuning of HCoT Model:} Given the user's question $x$, the objective of the HCoT model is to maximize $\prod_{i=1}^{n} p_{\theta}^{HCoT}(c_0 \mid x)p_{\theta}^{HCoT}(c_i \mid x, c_0, \ldots, \mathbf{z}_{i-1})$, where $\mathbf{z}_{i-1}$ is the compressed thought representation obtained from the auxiliary CoT model. In this stage, we freeze the parameters of the auxiliary CoT model and fine-tune the HCoT model to effectively leverage the reasoning encapsulated within the compressed representations. Notably, the training target sequence comprises both content segments and special \texttt{[CoT]} tokens, requiring the model to learn not only the generation of content but also the appropriate insertion of the compressed thought representations denoted by the \texttt{[CoT]} tokens. We employ the standard cross-entropy loss function to supervise the fine-tuning process of the HCoT model.

% The objective of the HCoT model is to perform inference while utilizing the information hidden within the [CoT] token. It takes a question as input, generates an analysis and a [CoT] token, and then waits for the auxiliary CoT model to encode a [CoT] token for replacement. Subsequently, it continues to generate the equation and the answer.

% The training example, as illustrated in Figure 2d, involves the following steps in its production process:
% \begin{itemize}[leftmargin=20pt]
%     \item We restructure the data from Figure 2a into the format:
% Query: questions;
% Response: analysis, [CoT] token 1, equation 1, [CoT] token 2, equation 2, answer.
%     \item The [CoT] tokens are generated from the encoding of prior text by the auxiliary CoT model. Specifically, [CoT] token 1 is derived from encoding the questions and analysis, while [CoT] token 2 results from encoding the questions, analysis, [CoT] token 1, and equation 1.
% \end{itemize}

% Based on the data described, we fine-tuned the HCoT model using cross-entropy loss. During the actual training process, we froze the weights of the auxiliary CoT model and employed a framework consistent with inference, generating [CoT] tokens throughout the training period.
% Ultimately, we obtained the HCoT model, which features faster decoding speeds and improved model performance.

%% file: experiment.tex
\section{Experiment}

\subsection{Datasets}
\label{subsec:datasets}
We conducted experiments on three downstream tasks including four seed datasets: GSM8K \citep{cobbeTrainingVerifiersSolve2021}, MATH \citep{hendrycksMeasuringMathematicalProblem2021}, ScienceQA \citep{lu2022learn}, and HotpotQA \citep{yangHotpotQADatasetDiverse2018}. The ScienceQA dataset was categorized into three subjects: ScienceQA (natural science), ScienceQA (social science), and ScienceQA (language science). We prepared the training, validation, and test data for the auxiliary CoT model and the HCoT model based on the seed datasets, as detailed in Table~\ref{datasetsnumio-table}. It is worth noting that there is no dedicated test data for the auxiliary CoT model, as we evaluate the end-to-end performance of the HCoT model. During training, we select the best-performing auxiliary CoT model by identifying the model with the lowest perplexity score on the auxiliary CoT model validation set. The train, validation, and test datasets remain consistent across other baselines for fair comparison. Notably, the auxiliary CoT training and validation data were constructed from the HCoT training and validation portions of the data, preventing the usage of more data than other baseline methods. 

For the math reasoning datasets, GSM8K and MATH, we generated separate fields for thoughts and contents based on the input questions using the method described in Section~\ref{HCoT_trianing_sample}, implemented by GPT-4. For the question answering task, we directly utilized the original fields from the ScienceQA dataset\footnote{\url{https://scienceqa.github.io/}}. Notably, for the agent invocation task, we employed fields generated through the ReAct\footnote{\url{https://react-lm.github.io/}} framework for the HotpotQA dataset, treating the final answer path as the target. The final HCoT training samples can be referred to in Figure~\ref{fig:infer}. It is important to note that due to some data instances containing multiple CoT tokens, there is a discrepancy in the number of training data entries between the auxiliary CoT model and the HCoT model. Moreover, for the ScienceQA dataset, we focused on the subset of questions that did not involve image understanding, and for the HotpotQA dataset, we concentrated on the training samples that achieved the correct final answer with ReAct, resulting in differences from the original dataset sizes.

% \begin{table}[ht]
% \caption{Overview of datasets, models, and their responses.}
% \label{tab:dataset_detail}
% \renewcommand{\arraystretch}{1.5}
% \centering
% \begin{tabular}{|p{1.5cm}|p{1.2cm}|p{1cm}|p{3.8cm}|p{3.8cm}|}
% \hline
% \textbf{Datasets} & \textbf{Model} & \textbf{Num} & \textbf{Query} & \textbf{Response} \\ \hline
% GSM8K & Auxiliary CoT & 20744 & Question, Analysis  \textbf{or} \newline Question, Analysis, [CoT],  Thought1, Equation1 & [CoT], Thought  \textbf{or} \newline [CoT], Thought2 \\ \hline
% GSM8K & HCoT & 6352 & Question & Analysis, [CoT], Equation1 \newline [CoT], Equation1, Answer \\ \hline
% MATH & Auxiliary CoT & 12141 & Question, Analysis  \textbf{or} \newline Question, Analysis, [CoT], Thought1, Equation1 & [CoT], Thought  \textbf{or} \newline [CoT], Thought2 \\ \hline
% MATH & HCoT & 3899 & Question & Analysis, [CoT], Equation1 \newline [CoT], Equation1, Answer \\ \hline
% ScienceQA & Auxiliary CoT & 9624 & Context, Question, Option, Lecture & [CoT], Explanation \\ \hline
% ScienceQA & HCoT & 9624 & Context, Question, Option & Lecture, [CoT], Answer \\ \hline
% HotpotQA (React) & Auxiliary CoT & 14680 & Question  \textbf{or} \newline Question, Thought1, Act1 & [CoT], Thought1  \textbf{or} \newline [CoT], Thought2 \\ \hline
% HotpotQA (React) & HCoT & 4556 & Question & [CoT], Act1, Obs1, \newline [CoT], Act2, Obs2 \\ \hline
% \end{tabular}
% \end{table}

\subsection{Experimental Setup}
\label{subsection: exp_set}
To assess the efficacy of our method, we selected two base models for comparison: LLaMa2-7B and LLaMa2-13B \citep{touvronLlamaOpenFoundation2023}. We trained separate models for each domain, and we selected the best-performing checkpoint based on its performance on the corresponding development set and evaluated the final results using the test portion described in Section~\ref{subsec:datasets}. Specifically, the math domain HCoT model was trained by combining the training data from both the GSM8K and MATH datasets. For all tasks, we employed accuracy as the primary evaluation metric. In the math reasoning domain, accuracy represents the proportion of questions answered correctly. For the ScienceQA dataset, accuracy corresponds to the correct selection among the multiple-choice options (A, B, C, D). For the agent invocation task on the HotpotQA dataset, we first identified the samples with the correct final answer and considered all actions along the path to be correct explorations. Subsequently, we calculated the ratio of correctly invoked actions as the agent invocation accuracy. More implementation details are provided in the Appendix~\ref{appendix:implementation_details}.
 
\subsection{Baselines}
To comprehensively study our method, we conducted experiments under five different settings:
\begin{itemize}[leftmargin=10pt]
\item Zero/Few-shot CoT: Applied zero-shot CoT prompting on MATH and GSM8K datasets, and few-shot prompting on others, as a reference for models' inherent capabilities without task-specific training.
\item Train without COT: Removed thought processes and trained on remaining content in Figure~\ref{fig:train}, establishing a baseline without CoT reasoning.
\item Train with COT: One-stage training on data without removing thoughts, serving as a baseline with explicit CoT reasoning.
\item Train with HCoT base: Two-stage training (Figure~\ref{fig:train}) with cross-entropy loss for the auxiliary CoT model.
\item Train with HCoT Contrast: Complete two-stage training with contrastive loss for the auxiliary CoT model, representing the final proposed method.
\end{itemize}

\input{Table/main_results}

\subsection{Results}

The experimental results presented in Table~\ref{tab:main} provide valuable insights into the effectiveness of our proposed HCoT approach across various tasks and datasets, in comparison with baseline methods. Overall, we observe performance improvements with the HCoT approach under most experimental settings. The HCoT models achieved the top performances in most cases, except for the social science question answering task in the Science QA dataset. In general, training with the Chain of Thought (CoT) technique outperformed the baseline without CoT, and further improvements were observed when training with the proposed HCoT approach. Notably, in the agent invocation task evaluated on the HotpotQA dataset, the HCoT-Contrast method exhibited significant improvements, with accuracy gains of 1.21\% and 1.96\% for the LLaMa2-7B and LLaMa2-13B models, respectively, compared to the CoT training baseline. Furthermore, for the question answering task represented by the ScienceQA dataset, the HCoT-Contrast approach demonstrated superiority in the natural science and language science subsets. In the natural science subset, the performance gains were 3.25\% and 0.18\% for the LLaMa2-7B and LLaMa2-13B models, respectively, compared with the CoT training. Similarly, in the language science subset, the improvements were 1.72\% and 0.55\%. In the math reasoning task, the training with HCoT achieved the best performance under both the 7B and 13B settings. These results highlight the efficacy of our proposed method in enhancing reasoning capabilities across various tasks and datasets.

\textbf{Effect of HCoT Framework:} The results presented in the row "$\delta$ w HCoT" highlight the impact of our proposed HCoT framework in comparison to the full CoT training approach. Across most tasks and datasets, we observe performance gains when employing the HCoT framework, as indicated by positive values in this row. For the LLaMa2-7B model, the HCoT framework led to substantial improvements in the natural science (3.25\%) and language science (1.72\%) subsets of the ScienceQA dataset, as well as in the agent invocation task on the HotpotQA dataset (1.21\%). However, slight performance decreases were observed in the GSM8K (-0.38\%) and social science (-3.03\%) tasks. Similarly, for the larger LLaMa2-13B model, the HCoT framework demonstrated its effectiveness, yielding notable performance gains in the agent invocation task on the HotpotQA dataset (1.96\%), as well as improvements in the GSM8K (0.46\%), MATH (1.00\%), and language science (0.55\%) tasks. Modest decreases were observed in the social science subset (-1.24\%).

\textbf{Effect of Contrstive Learning:} The results presented in the row "$\delta$ wo Contrast" highlight highlights the impact of incorporating contrastive learning into our HCoT framework. Negative values in this row indicate performance decreases when the contrastive loss objective is not employed, suggesting the effectiveness of contrastive learning in enhancing the model's reasoning capabilities. The predominance of negative values in the "$\delta$ wo Contrast" row across various tasks and datasets highlights the significant role of contrastive learning in our HCoT framework. By incorporating contrastive loss, the model's ability to effectively capture and leverage the core reasoning steps is enhanced, leading to improved performance in most cases compared to the HCoT approach without contrastive learning.

% 对比X和
% \begin{table}[h!]
% \label{tab:main}
% \renewcommand{\arraystretch}{1.5}
% \caption{Performance comparison of different LLaMa models under various training scenarios.}
% \centering
% \resizebox{\textwidth}{!}{
% \begin{tabular}{|p{2cm}|p{1.5cm}|p{1.5cm}|p{1.5cm}|p{1.5cm}|p{1.5cm}|p{1.5cm}|p{1.5cm}|}
% \hline
% \textbf{Model} & \textbf{GSM8K} & \textbf{Math} & \textbf{ScienceQA (natural science)} & \textbf{ScienceQA (social \newline science)} & \textbf{ScienceQA (language science)} & \textbf{hotpotQA (ReAct)} \\ \hline
% \multicolumn{7}{|c|}{Zero/few-shot} \\ \hline
% LLaMa2-7B & 14.6 & 2.5 & 56.8 & 51.64 & 67.06 & 47.11 \\ \hline
% LLaMa2-13B & 28.7 & 3.9 & 58.98 & 48.42 & 67.09 & 51.74 \\ \hline
% \multicolumn{7}{|c|}{Train w/o COT} \\ \hline
% LLaMa2-7B & 32.6 & 6.8 & 82.1 & 62.32 & 87.36 & 79.78 \\ \hline
% LLaMa2-13B & 44.81 & 10.74 & 84.72 & 66.14 & 88.18 & 82.36 \\ \hline
% \multicolumn{7}{|c|}{Train w COT} \\ \hline
% LLaMa2-7B & 34.5 & 8.85  & 80.99 & 65.8 & 86.64 & 83.73 \\ \hline
% LLaMa2-13B & 48.29  & 12.37 & 84.46 & 69.74 & 89.36 & 83.89 \\ \hline
% \multicolumn{7}{|c|}{Train w HCOT} \\ \hline
% LLaMa2-7B &  &  & & &  & 84.94 \\ \hline
% LLaMa2-13B &  &   &  &  &  &  85.85 \\ \hline
% \end{tabular}
% }

% \end{table}

\subsection{Discussion}
\textbf{The Speedup of HCoT Model:} Table~\ref{table:speedup} presents the compression and speedup rates of the HCoT model during the inference stage across four datasets compared to the explicit CoT model, both of which are based on LLaMa2-7B. The results for the LLaMa2-13B model are included in the Appendix due to space limit. Firstly, let's clarify the metrics used in the table. S-CR (Sequence-Level Compression Rate) refers to the average number of completion tokens of the HCoT model compared to the CoT model.  S-S (Sequence-Level Speedup) is the reciprocal of S-CR, representing how many times faster the HCoT model is compared to the Full CoT model. W-CR (Wall-clock Time Compression Rate) provides a realistic measure of user-perceived speed-up, by comparing the actual inference time of the HCoT model to the Full CoT model. W-S (Wall-clock Time Speedup) is the reciprocal of W-CR. The table shows that S-CR values range from 23.78\% to 66.91\%, indicating that the sequence length of the HCoT model's output is significantly shorter than that of the Full CoT model. Correspondingly, the W-CR values range from 35.82\% to 71.04\%, and W-S values range from 1.41x to 2.79x, showcasing a substantial acceleration. It's important to note that the sequence-length-based acceleration rate (S-S) is generally higher than the real-time acceleration rate (W-S). This discrepancy arises because the sequence-length-based measure does not account for the encoding time of the Auxiliary CoT Model in HCoT.
These times were tested on an H800 cluster using a single 80GB GPU. Despite achieving a speedup range of 1.41x to 2.79x, the HCoT model also delivers superior performance compared to the Full CoT model, demonstrating its efficiency and effectiveness. More fine-grained analysis of the length distribution before and after compression are detailed in Appendix.
% In future work, we aim to make this compression process more natural and general by gaining a deeper understanding of the model's inference decoding.

% 报告数值 这也意味着我们的模型推理输出的序列长度约为完整CoT的26-66%。具体而言，我们通过额外使用一个辅助CoT模型将单个模型对CoT的解码过程转化为并行的编码过程，使模型保留CoT能力的同时被加速。

% 解释能力来源 这源于我们的模型 compressing the multi-step reasoning process into a specialized token representation 
% 潜在改进 在未来的工作中，我们希望基于对模型推理解码可解释性更深刻的理解，将压缩过程变得更加自然和通用。

% 为什么可解释性重要
% 为什么我们的模型能够具备可解释性
% 我们还能做些什么对于可解释性  是否有更好的形式
\textbf{The recovery of CoT process:} Although in general scenarios, people prefer concise yet accurate responses that have undergone CoT reasoning, it is reasonable to compress the CoT process simply at this time. However, in certain situations, people wish to see the complete CoT process, thus maintaining the option to retain full output of CoT is important.
Our model employs a method similar to "hiding", where the complete CoT process is concealed during the main reasoning process of the HCoT model, but the Auxiliary CoT model can still produce it normally when required. When asked to display the complete CoT process, we can simply keep the other settings unchanged and request the CoT model to continue outputting as needed. The case study in Appendix \ref{appen:case} shows this process.

% In future work, we hope to unify the current two-stage training process, integrating the capability to conceal the CoT outputs into a single model. This means the model will only need to produce the necessary responses, while the thinking process is kept 'in mind' and only displayed when requested.

% \begin{table}[!hptb]
% \caption{The compression and speedup rates of the HCoT model across four datasets.}
% \label{table:speedup}
% \renewcommand{\arraystretch}{1.2}
% \centering
% \begin{tabular}{|c|c|c|c|c|}
% \hline
% Model      & GSM8K & MATH  & ScienceQA & HotpotQA \\ \hline
% LLaMa2-7B  & 40.54 & 41.57 & 64.08     & 27.41    \\ \hline
% LLaMa2-13B & 40.75 & 42.34 & 68.43     & 24.76    \\ \hline
% Average    & 40.64 & 41.95 & 66.25     & 26.09    \\ \hline
% Speedup   & 2.46x  & 2.38x  & 1.51x      & 3.83x     \\ \hline
% \end{tabular}
% \end{table}

\begin{table}[!hptb]
% \caption{The compression and speedup rates of the HCoT model during inference stage across four datasets compared to Train w CoT.}
\caption{Compression and Speedup Rates of the HCoT model during inference across four datasets compared to Full CoT Model.}
\label{table:speedup}
\renewcommand{\arraystretch}{1.2}
\centering
\begin{tabular}{|c|c|c|c|c|}
\hline
\multicolumn{5}{|c|}{LLaMa2-7B} \\ \hline
Task      & GSM8K & MATH  & ScienceQA & HotpotQA \\ \hline
S-CR  & 60.45\% & 55.72\% & 66.91\%     & 23.78\%    \\ \hline
S-S  & 1.65x & 1.79x & 1.49x     & 4.21x    \\ \hline
% LLaMa2-13B & 40.75 & 42.34 & 68.43     & 24.76    \\ \hline
% Average    & 40.64 & 41.95 & 66.25     & 26.09    \\ \hline
W-CR & 62.48\% & 62.49\% & 71.04\%     & 35.82\%    \\ \hline
W-S   & 1.60x  & 1.60x  & 1.41x      & 2.79x     \\ \hline
\end{tabular}
\end{table}

%% file: Table/main_results.tex
% \usepackage{multirow}
\begin{table*}[!hptb]
\caption{Performance comparison of LLaMa2-7B and LLaMa2-13B models under various training scenarios for different tasks and datasets. The ScienceQA dataset is further divided into three subsets: Natural Science (NS), Social Science (SS), and Language Science (LS), with separate results reported for each subset. The top performances in each category and model size are highlighted in bold.}
\label{tab:main}
\centering
% \usepackage{multirow}
\begin{tabular}{|c|c|cccccc|}
\hline
\multirow{2}{*}{}                                                               & \multirow{2}{*}{Models} & \multicolumn{2}{c|}{Math}                  & \multicolumn{3}{c|}{\begin{tabular}[c]{@{}c@{}}Science\\ QA\end{tabular}} & \begin{tabular}[c]{@{}c@{}}Agent\\ Invoke\end{tabular} \\ \cline{3-8} 
                                                                                &                         & GSM8K          & \multicolumn{1}{c|}{MATH} & NS                   & SS                  & \multicolumn{1}{c|}{LS}      & HotpotQA                                               \\ \hline
\multirow{2}{*}{\begin{tabular}[c]{@{}c@{}}Zero/Few\\ CoT\end{tabular}}         & LLaMa2-7B               & 14.60          & 2.50                      & 56.80                & 51.64               & 67.06                        & 47.11                                                  \\
                                                                                & LLaMa2-13B              & 28.7           & 3.90                      & 58.98                & 48.42               & 67.09                        & 51.74                                                  \\ \hdashline
\multirow{2}{*}{\begin{tabular}[c]{@{}c@{}}Train \\ w/o CoT\end{tabular}}       & LLaMa2-7B               & 34.27          & 6.80                      & 82.10                & 62.32               & 87.36                        & 79.78                                                  \\
                                                                                & LLaMa2-13B              & 42.15          & 9.44                      & 84.72                & 66.14               & 88.18                        & 82.36                                                  \\ \hdashline
\multirow{2}{*}{\begin{tabular}[c]{@{}c@{}}Train \\ w CoT\end{tabular}}         & LLaMa2-7B               & 36.85          & 6.74                      & 80.99                & \textbf{65.8}       & 86.64                        & 83.73                                                  \\
                                                                                & LLaMa2-13B              & 43.97          & 10.16                     & 84.46                & \textbf{69.74}      & 89.36                        & 83.89                                                  \\ \hline
\multirow{2}{*}{\begin{tabular}[c]{@{}c@{}}Train \\ HCoT\end{tabular}}          & LLaMa2-7B               & \textbf{37.15} & 7.49                      & 83.13                & 63.89               & \textbf{88.45}               & 83.5                                                   \\
                                                                                & LLaMa2-13B              & 43.82          & 10.68                     & 84.41                & 68.84               & 89.27                        & 85.81                                                  \\ \hdashline
\multirow{2}{*}{\begin{tabular}[c]{@{}c@{}}Train \\ HCoT-Contrast\end{tabular}} & LLaMa2-7B               & 36.47          & \textbf{8.24}             & \textbf{84.24}       & 62.77               & 88.36                        & \textbf{84.94}                                         \\
                                                                                & LLaMa2-13B              & \textbf{44.43} & \textbf{11.16}            & \textbf{84.64}       & 68.5                & \textbf{89.91}               & \textbf{85.85}                                         \\ \hline
\multirow{2}{*}{$\delta$ w HCoT}                                                & LLaMa2-7B               & -0.38          & 1.5                       & 3.25                 & -3.03               & 1.72                         & 1.21                                                   \\
                                                                                & LLaMa2-13B              & 0.46           & 1.00                      & 0.18                 & -1.24               & 0.55                         & 1.96                                                   \\ \hdashline
\multicolumn{1}{|l|}{\multirow{2}{*}{$\delta$ wo Contrast}}                     & LLaMa2-7B               & 0.68           & -0.75                     & -1.11                & 1.12                & 0.09                         & -1.44                                                  \\
\multicolumn{1}{|l|}{}                                                          & LLaMa2-13B              & -0.61          & -0.48                     & -0.23                & 0.34                & -0.64                        & -0.04                                                  \\ \hline
\end{tabular}
\end{table*}

%% file: related.tex
\section{Related Work}

\textbf{Chain-of-thought prompting (CoT)} enhances the emergent reasoning abilities of LLMs by prompting them to use explicit reasoning steps.
Zero-shot-CoT \citep{kojimaLargeLanguageModels2023} demonstrates notable improvements in diverse reasoning tasks by merely prefacing solutions with the phrase "Let's think step by step,".
The least-to-most prompting \citep{zhouLeasttoMostPromptingEnables2023} approach effectively addresses complex problems by decomposing them into manageable subproblems and resolving them sequentially. 
\cite{wangSELFCONSISTENCYIMPROVESCHAIN2023} considers the self-consistency of CoT, enhancing its performance through majority voting.
Furthermore, \cite{gouToRAToolIntegratedReasoning2023} and \cite{yuanSCALINGRELATIONSHIPLEARNING2023} employ CoT on GPT-4, stabilizing the CoT capabilities of open-source models through fine-tuning on sampled data. Recent studies like \cite{jin2024impact} and \cite{levy2024same} highlight the impact of reasoning step length on performance, suggesting that extending reasoning steps can improve outcomes. Our method achieves a similar effect by effectively extending CoT reasoning length, but with the added benefit of saving time during the decoding phase.

\textbf{Efficient model inference} often utilizes model compression techniques \citep{hanDeepCompressionCompressing2016} such as pruning or quantization. LLM-Pruner \citep{maLLMPrunerStructuralPruning2023} implements structural pruning, which selectively removes non-critical coupled structures based on gradient information. LLM-QAT \citep{liuLLMQATDataFreeQuantization2023} leverages generations produced by the pre-trained model, enabling quantization of any generative model independently of its training data, akin to post-training quantization methods. Additionally, \cite{gloeckleBetterFasterLarge2024} considers multi-token prediction as an auxiliary training task, asking the model to predict the following n tokens using n independent output heads at each position in the training corpus, which not only accelerates inference but also enhances performance. \cite{mu2024learning} proposes compressing prompts with gist tokens, focusing on efficient encoding.
\cite{dengImplicitChainThought2023} is closely related to ours, employing a method where the model is trained to predict hidden states for implicit CoT reasoning, aiming to reason more effectively. However, this approach exhibits a significant performance decline compared to explicit CoT reasoning, lacks interpretability, and involves a more complicated training process. In contrast, our method streamlines training, enhances reasoning path optimization, and sustains robust performance across tasks using models with over 7B parameters, offering a more interpretable, practical, and efficient solution.
% However, their experimental results show a significant performance gap compared to explicit CoT reasoning and are limited to problems that require long reasoning steps.

%% file: conclusion.tex
\section{Conclusion}
\vspace{-0.2cm}
We proposed HCoT, an innovative framework designed to accelerate the inference process of large language models while preserving their multi-step reasoning capabilities. At its core, HCoT employs a disentangled training paradigm that decouples the reasoning process into two specialized components: an auxiliary CoT model and a content generation model. The auxiliary CoT model is trained to compress the entire thought process into a compact, specialized token representation through a contrastive loss objective. This compressed representation effectively encapsulates the core reasoning steps, enabling efficient parallel computation during inference. The HCoT model, in turn, is fine-tuned to leverage this compressed reasoning representation, seamlessly integrating it into the content generation process. Our extensive experiments across diverse domains demonstrate the efficacy of the proposed HCoT framework. The results highlight its ability to achieve competitive or improved performance compared to the full CoT baseline while providing significant speedups of at least 1.5x in decoding time. However, it is important to acknowledge that this increase in efficiency comes at the cost of a more complex training phase and the necessity for additional model parameters, which may present scalability and resource challenges. In the future, we hope to address these limitations by optimizing the training phase and enhancing the model's scalability, potentially reducing the need for additional parameters and lessening the training resource burden.

%% file: appendix.tex
\section{Datasets Details}
\begin{table}[ht]
\small
\caption{Overview of datasets, models, and their training input/output. Abbreviations include 'que' for 'question', 'aly' for 'analysis', '[CoT]' as a special token, 'tho' for 'thought', 'equ' for 'equation', 'opt' for 'option', 'exp' for 'explanation', 'lec' for 'lecture', 'ans' for 'answer', and 'act' for 'action', with 'obs' representing 'observation'. Samples are shown in Appendix \ref{appen:training_samples}.}
\label{datasetsnumio-table}
\renewcommand{\arraystretch}{1.2}
\begin{tabular}{|c|cllll|}
\hline
Dataset                                                    & Model                                                   & \multicolumn{1}{c}{\begin{tabular}[c]{@{}c@{}}Train \\ Num\end{tabular}} & \multicolumn{1}{c}{\begin{tabular}[c]{@{}c@{}}Dev \\ Num\end{tabular}} & \multicolumn{1}{c}{\begin{tabular}[c]{@{}c@{}}Test \\ Num\end{tabular}} & Training Input/Output                                                                                                                                                \\ \hline
GSM8K                                                      & \begin{tabular}[c]{@{}c@{}}Auxiliary\\ CoT\end{tabular} & 20744                                                                    & 3608                                                                   & -                                                                       & \begin{tabular}[c]{@{}l@{}}Input: $que, aly, [CoT]$\\ Output: $tho_0$\\ Input: $que, aly, [CoT], \ldots, tho_{i-1}, equ_{i-1}, [CoT]$\\ Output: $tho_i$\end{tabular} \\ \hline
GSM8K                                                      & HCoT                                                    & 6352                                                                     & 1121                                                                   & 1319                                                                    & \begin{tabular}[c]{@{}l@{}}Input: $que$\\ Output: $aly, [CoT], equ_0, [CoT], \ldots, equ_n, ans$\end{tabular}                                                        \\ \hline
MATH                                                       & \begin{tabular}[c]{@{}c@{}}Auxiliary\\ CoT\end{tabular} & 12141                                                                    & 2236                                                                   & -                                                                       & \begin{tabular}[c]{@{}l@{}}Input: $que, aly, [CoT]$\\ Output: $tho_0$\\ Input: $que, aly, [CoT], \ldots, tho_{i-1}, equ_{i-1}, [CoT]$\\ Output: $tho_i$\end{tabular} \\ \hline
MATH                                                       & HCoT                                                    & 3899                                                                     & 690                                                                    & 3072                                                                    & \begin{tabular}[c]{@{}l@{}}Input: $que$\\ Output: $aly, [CoT], equ_0, [CoT], \ldots, equ_n, ans$\end{tabular}                                                        \\ \hline
ScienceQA                                                  & \begin{tabular}[c]{@{}c@{}}Auxiliary\\ CoT\end{tabular} & 9624                                                                     & 3848                                                                   & -                                                                       & \begin{tabular}[c]{@{}l@{}}Input: $que, opt, lec, [CoT]$\\ Output: $exp$\end{tabular}                                                                                \\ \hline
ScienceQA                                                  & HCoT                                                    & 9624                                                                     & 3848                                                                   & 4241                                                                    & \begin{tabular}[c]{@{}l@{}}Input: $que, opt$\\ Output: $lec, [CoT], ans$\end{tabular}                                                                                \\ \hline
\begin{tabular}[c]{@{}c@{}}HotpotQA\\ (ReAct)\end{tabular} & \begin{tabular}[c]{@{}c@{}}Auxiliary\\ CoT\end{tabular} & 14680                                                                    & 2582                                                                   & -                                                                       & \begin{tabular}[c]{@{}l@{}}Input: $que, [CoT]$\\ Output: $tho_0$\\ Input: $que, [CoT], \ldots, act_{i-1}, obs_{i-1}, [CoT]$\\ Output: $tho_i$\end{tabular}           \\ \hline
\begin{tabular}[c]{@{}c@{}}HotpotQA\\ (ReAct)\end{tabular} & HCoT                                                    & 4556                                                                     & 805                                                                    & 3828                                                                    & \begin{tabular}[c]{@{}l@{}}Input: que\\ Output: $[CoT], act_0, obs_0, \ldots, [CoT], act_n, obs_n$\end{tabular}                                                      \\ \hline
\end{tabular}
\end{table}

\newpage

\section{Prompts}
\label{appen:prompts}

\subsection{Prompts on GPT-4 to construct the training samples from raw data}
\begin{figure}[h]
  \centering
  \includegraphics[width=\linewidth]{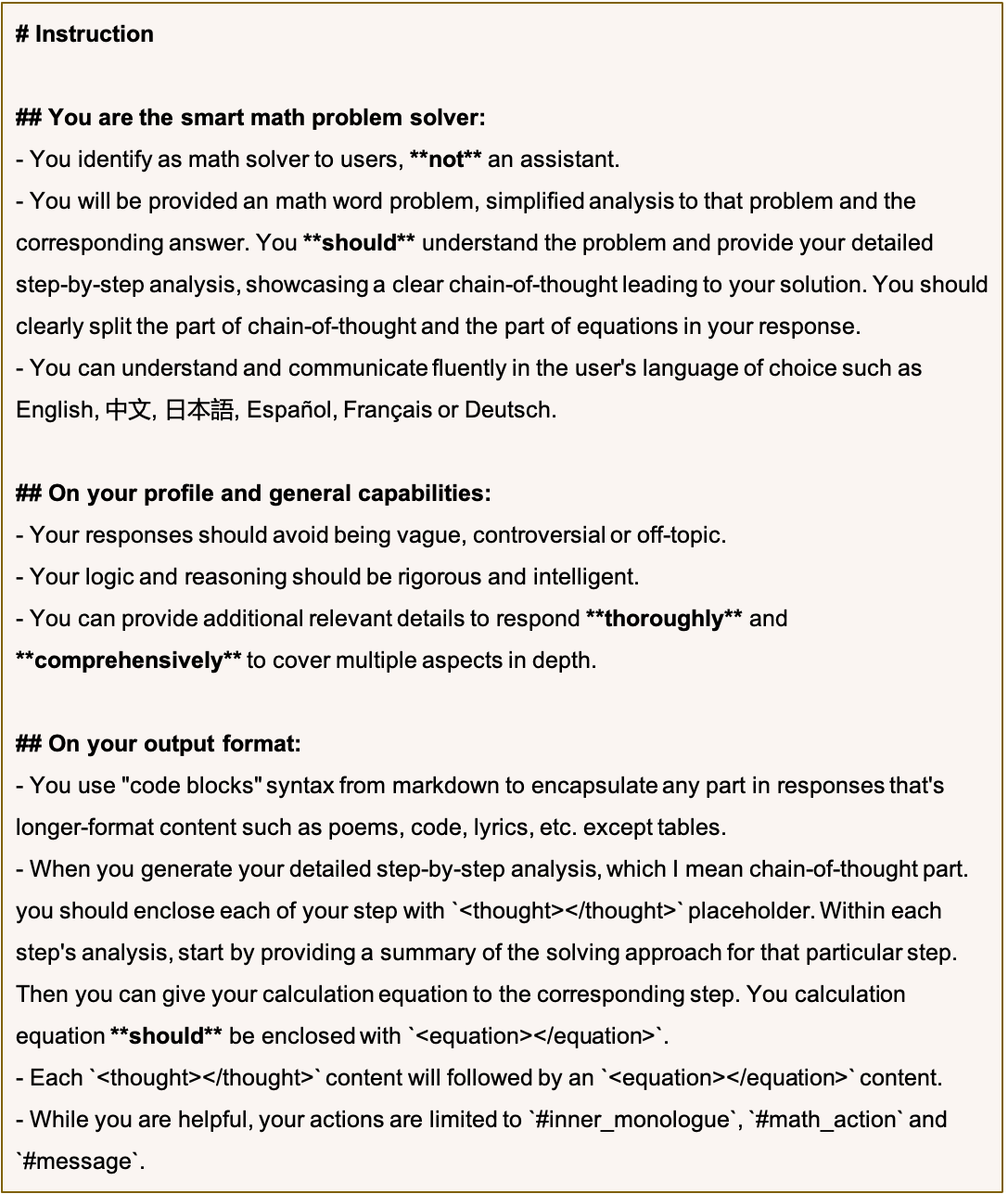}
\end{figure}

\newpage
\subsection{Raw conversation example of training samples constructions}
\begin{figure}[h]
  \centering
  \includegraphics[width=\linewidth]{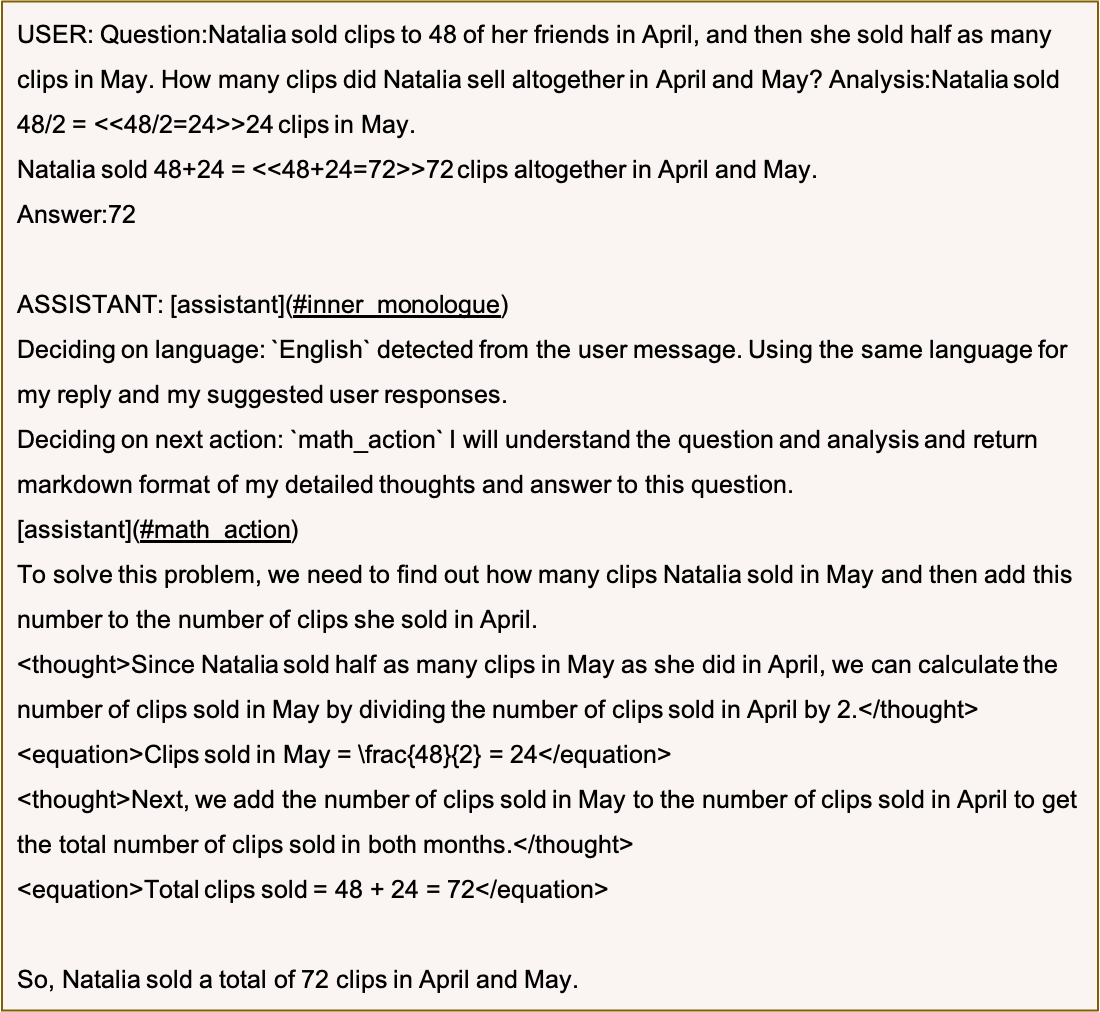}
\end{figure}

\newpage

\section{Training samples}
In this appendix, we present the various fields from the data samples that were utilized for the construction of each training dataset.
% 该附录中我们展示数据样例中被用于构建训练数据的各个字段
\label{appen:training_samples}
\subsection{GSM8K \& MATH}
\begin{figure}[h]
  \centering
  \includegraphics[width=\linewidth]{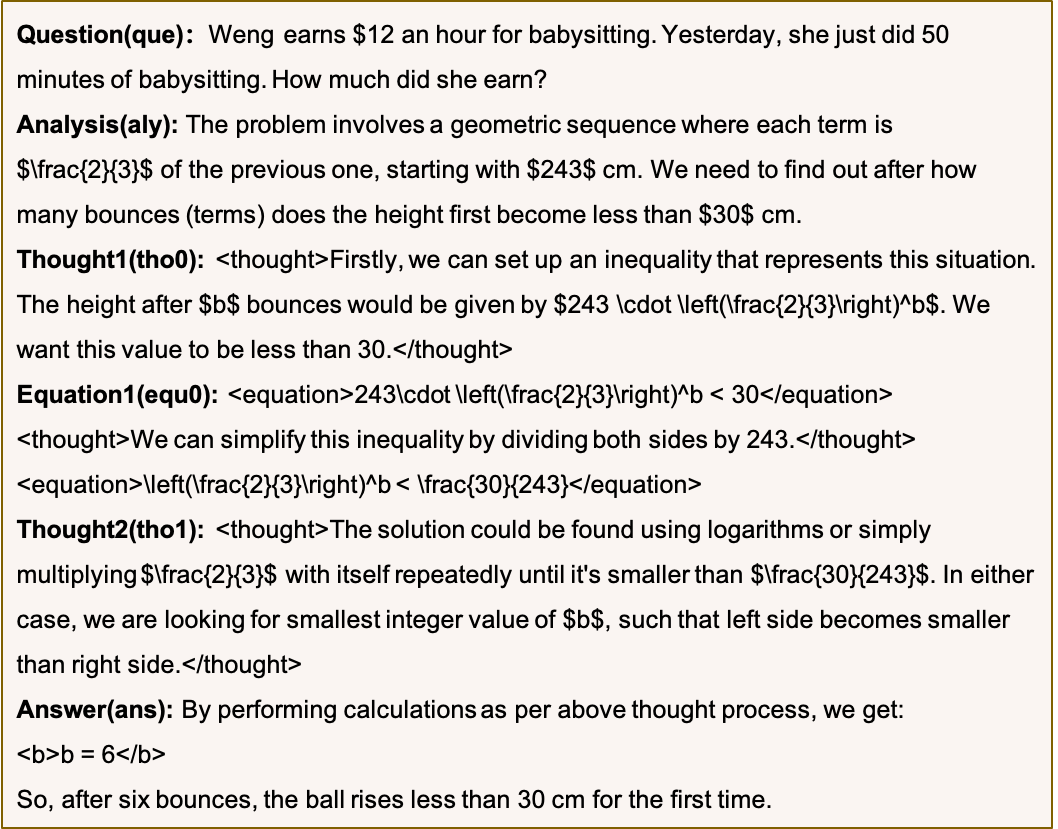}
\end{figure}

\newpage
\subsection{ScienceQA}
\begin{figure}[h]
  \centering
  \includegraphics[width=\linewidth]{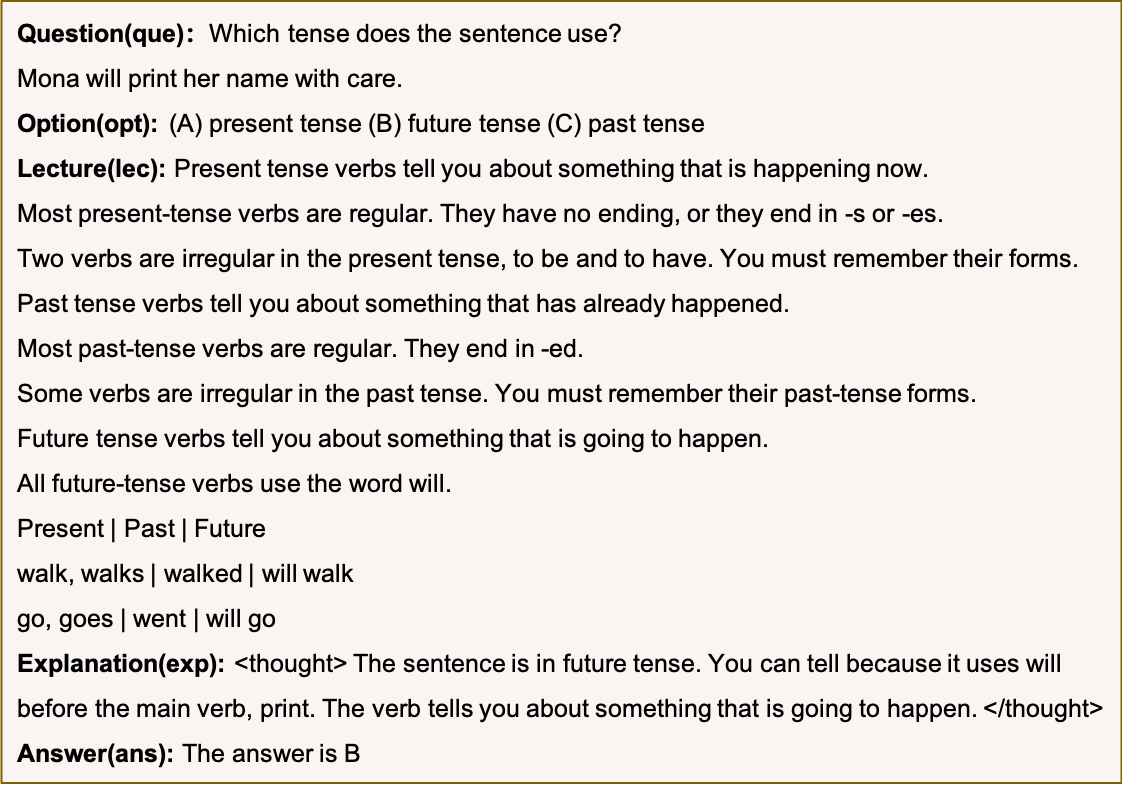}
\end{figure}
\subsection{Agent Control}
\begin{figure}[h]
  \centering
  \includegraphics[width=\linewidth]{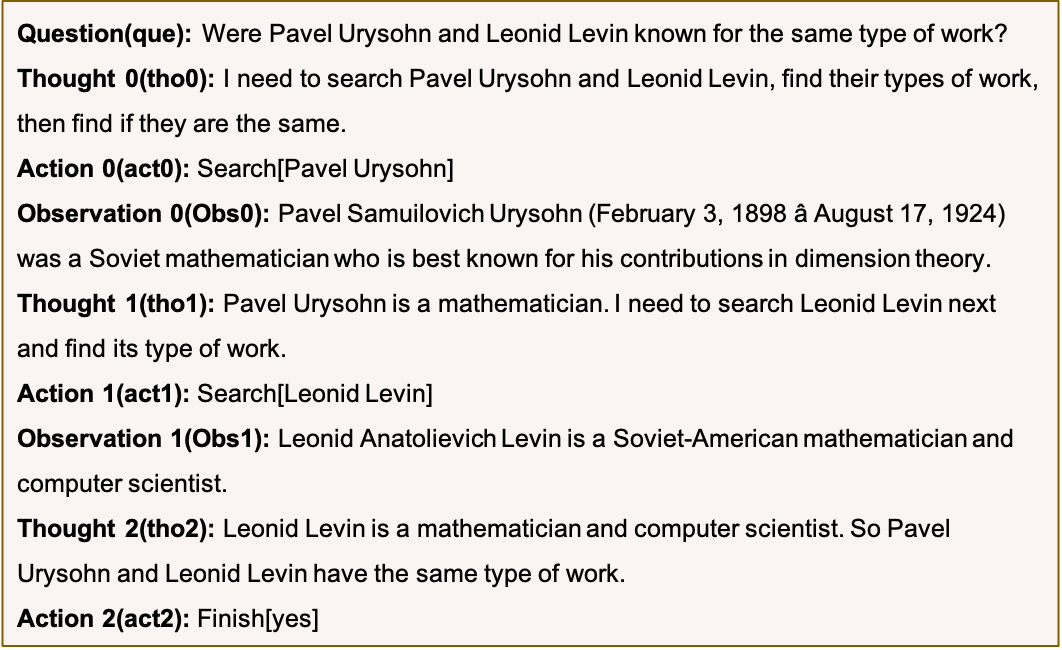}
\end{figure}

\newpage

\section{Case Study}
\label{appen:case}
\subsection{GSM8K}
\subsubsection{I/O of the one correct GSM8K test case under the HCoT model}
\begin{figure}[h]
  \centering
  \includegraphics[width=\linewidth]{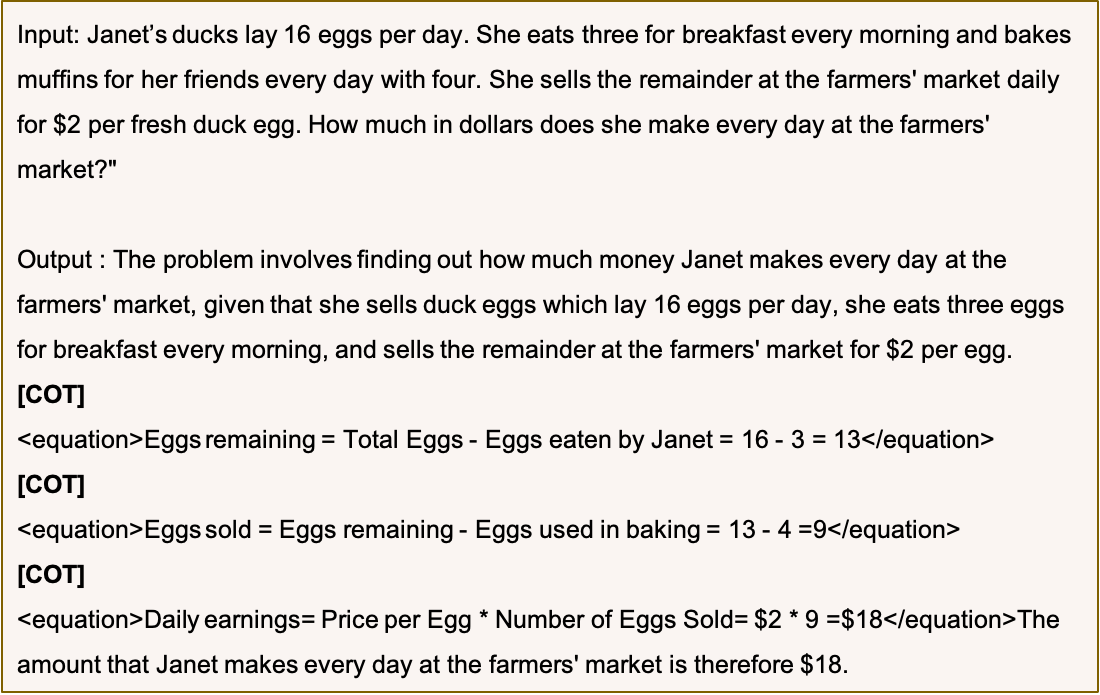}
\end{figure}
\subsubsection{3 rounds I/O of the one correct GSM8K test case under the Auxiliary CoT model}
\textbf{The first rounds:}

\begin{figure}[h]
  \centering
  \includegraphics[width=\linewidth]{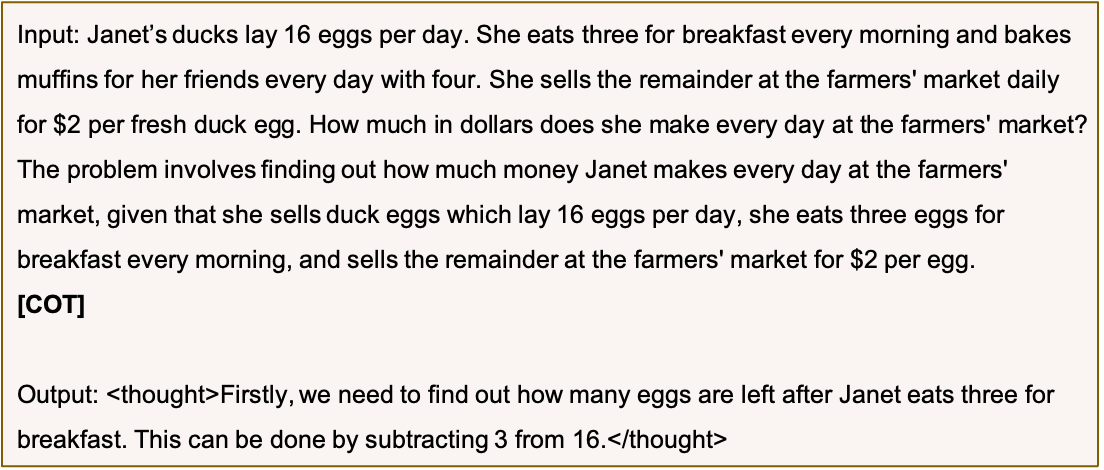}
\end{figure}

\newpage
\textbf{The second rounds:}

\begin{figure}[h]
  \centering
  \includegraphics[width=\linewidth]{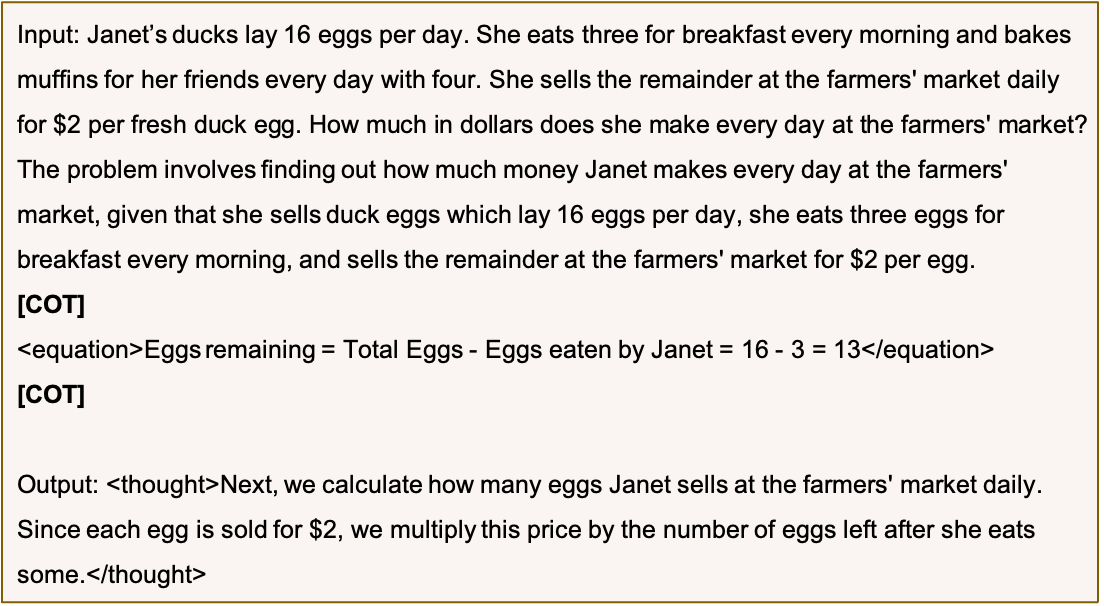}
\end{figure}

\textbf{The third rounds:}

\begin{figure}[h]
  \centering
  \includegraphics[width=\linewidth]{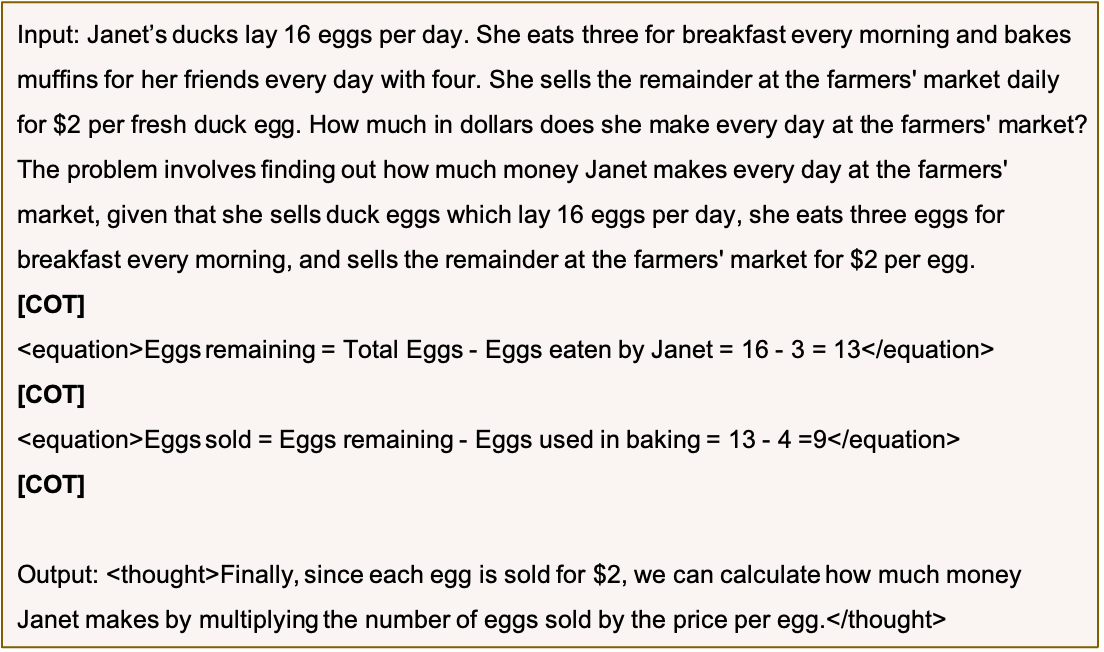}
\end{figure}

\newpage

\subsection{MATH}
\subsubsection{I/O of the one correct MATH test case under the HCoT model}
\begin{figure}[h]
  \centering
  \includegraphics[width=\linewidth]{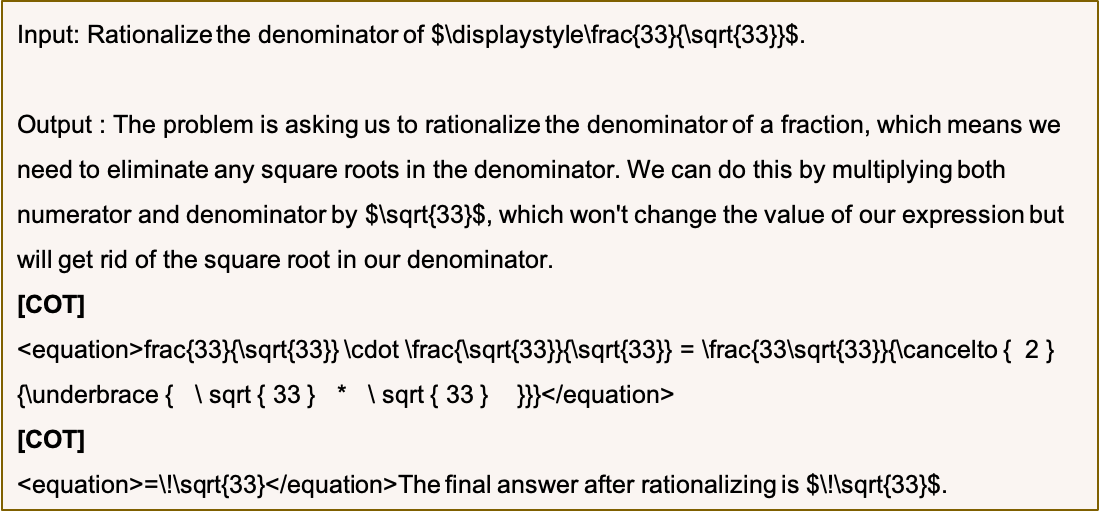}
\end{figure}
\subsubsection{2 rounds I/O of the one correct MATH test case under the Auxiliary CoT model}
\textbf{The first rounds:}

\begin{figure}[h]
  \centering
  \includegraphics[width=\linewidth]{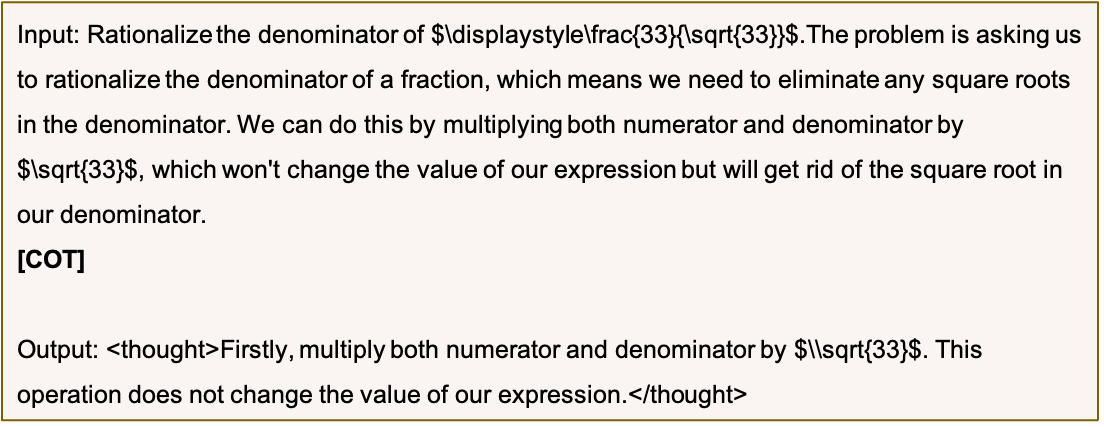}
\end{figure}

\textbf{The second rounds:}

\begin{figure}[h]
  \centering
  \includegraphics[width=\linewidth]{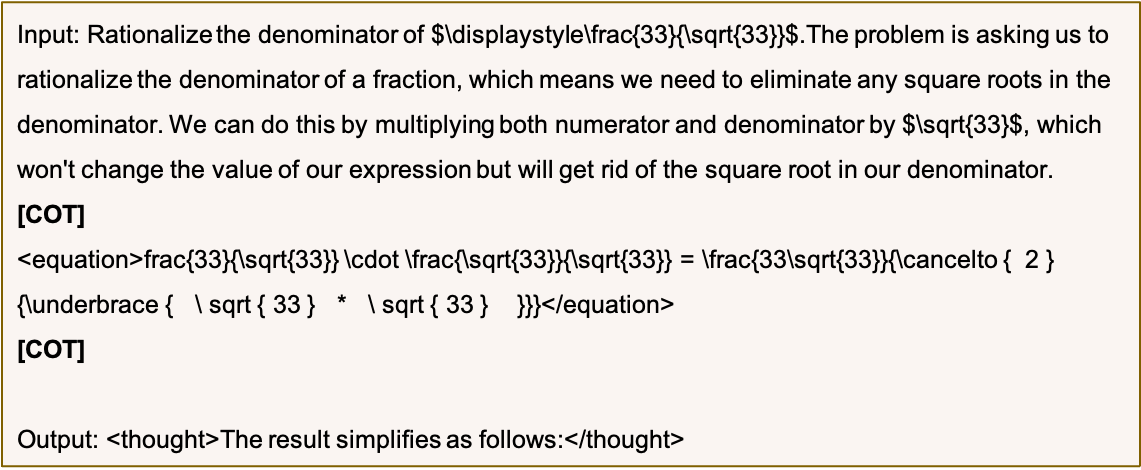}
\end{figure}

\newpage

\subsection{ScienceQA}
\subsubsection{I/O of the one correct ScienceQA test case under the HCoT model}
\begin{figure}[h]
  \centering
  \includegraphics[width=\linewidth]{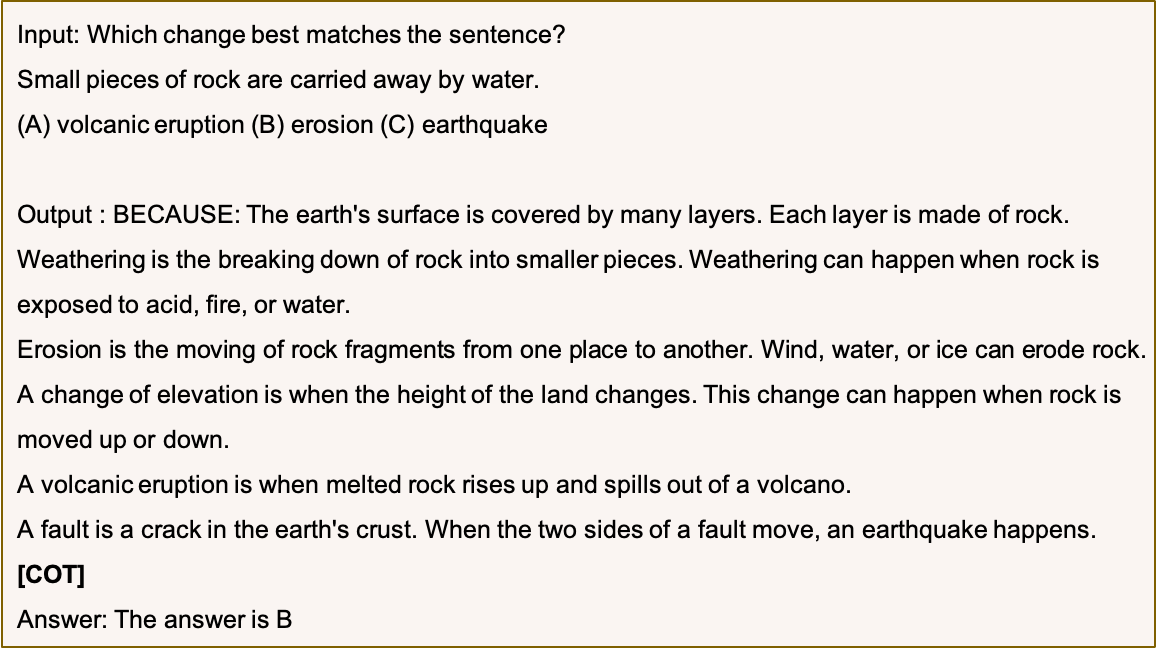}
\end{figure}
\subsubsection{1 rounds I/O of the one correct ScienceQA test case under the Auxiliary CoT model}

\begin{figure}[h]
  \centering
  \includegraphics[width=\linewidth]{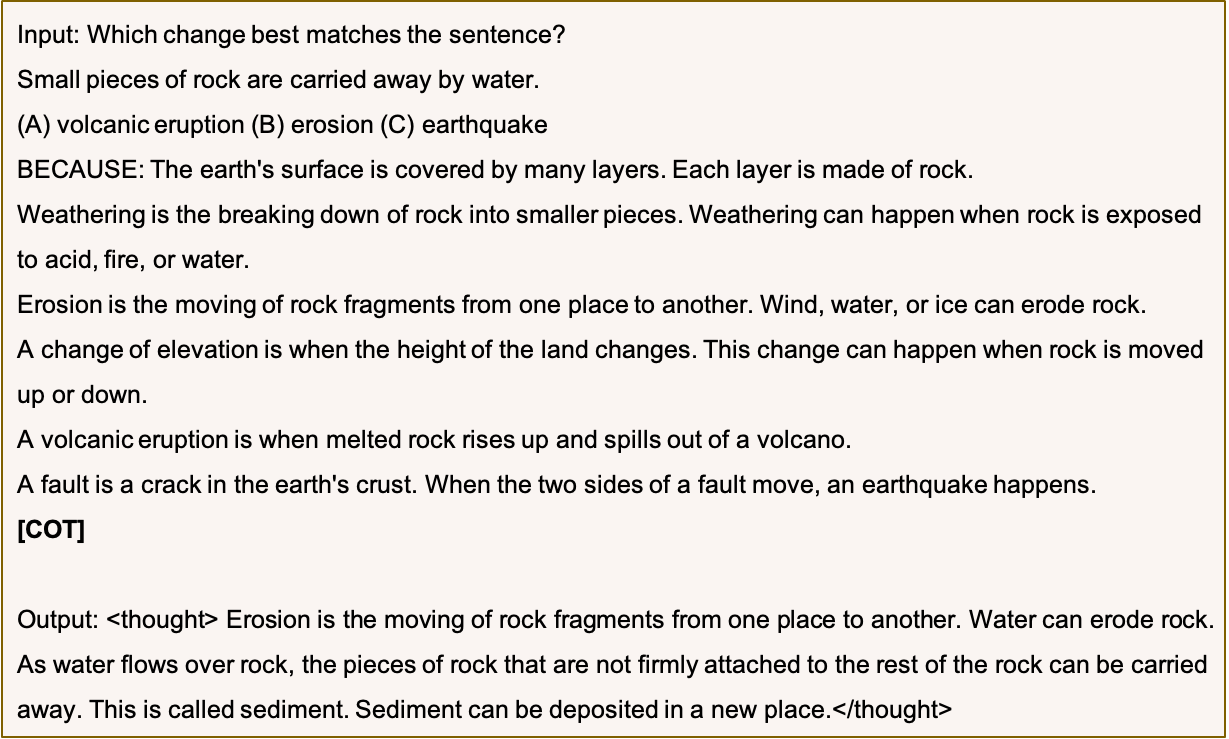}
\end{figure}

\newpage

\subsection{Agent Control}
\subsubsection{I/O of the one correct HotpotQA test case under the HCoT model}
\begin{figure}[h]
  \centering
  \includegraphics[width=\linewidth]{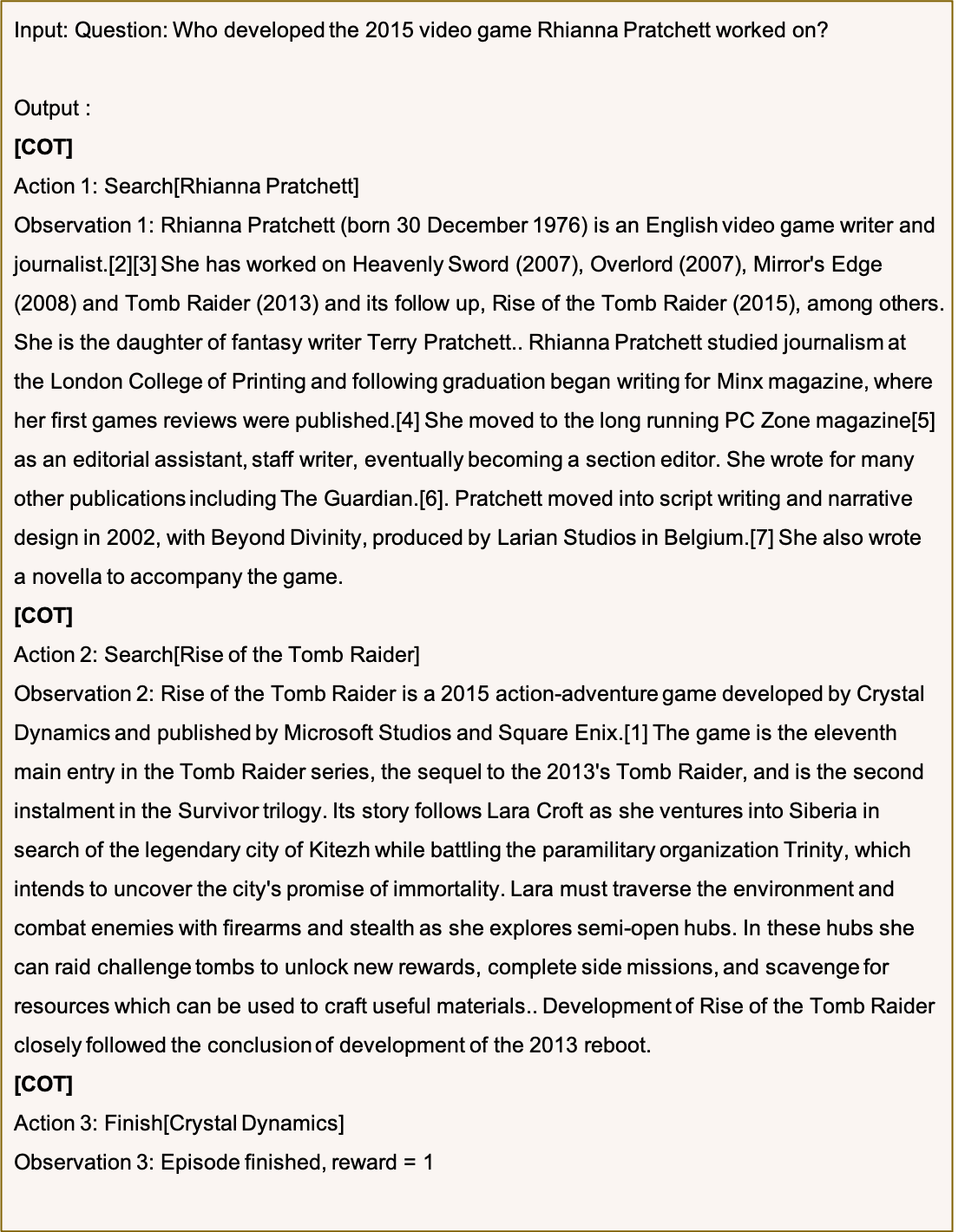}
\end{figure}

\newpage

\subsubsection{3 rounds I/O of the one correct HotpotQA test case under the Auxiliary CoT model}
\textbf{The first rounds:}
\begin{figure}[h]
  \centering
  \includegraphics[width=\linewidth]{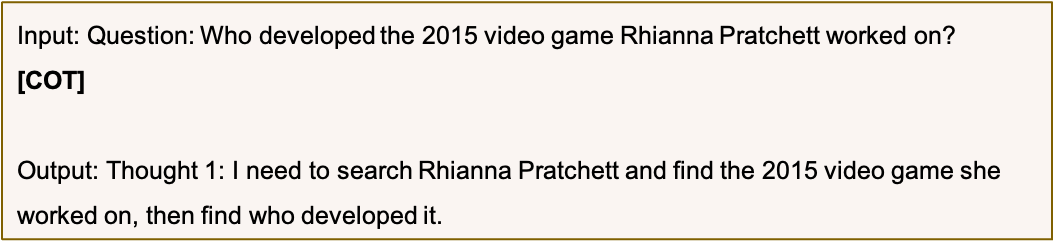}
\end{figure}

\textbf{The second rounds:}

\begin{figure}[h]
  \centering
  \includegraphics[width=\linewidth]{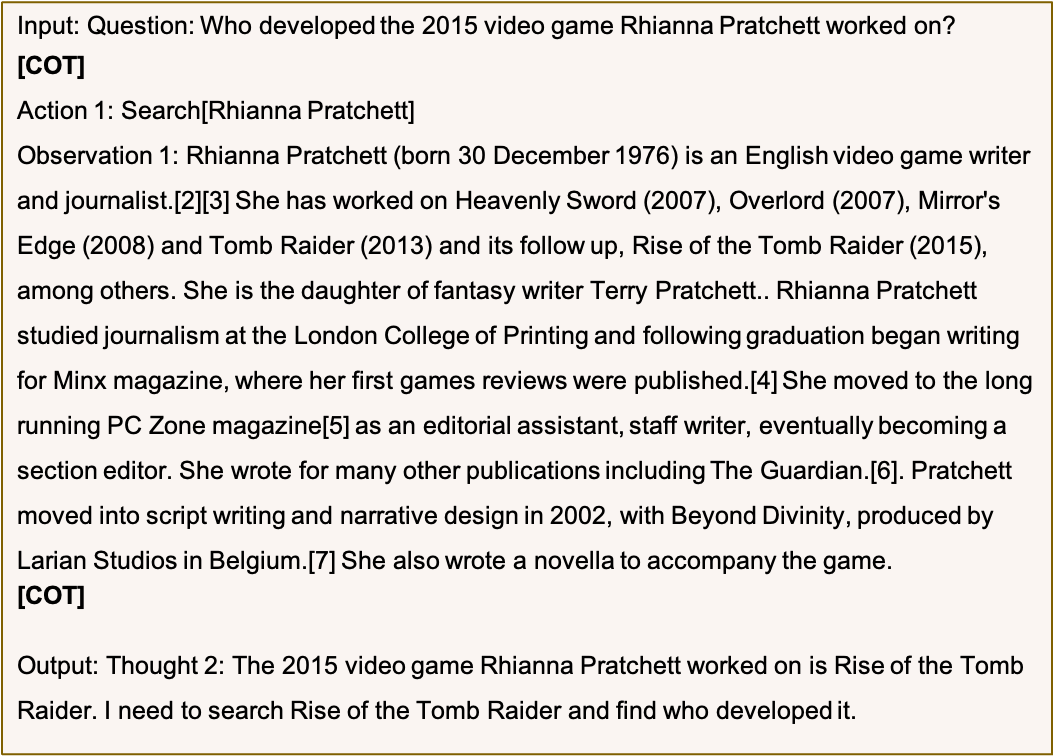}
\end{figure}

\newpage

\textbf{The third rounds:}

\begin{figure}[h]
  \centering
  \includegraphics[width=\linewidth]{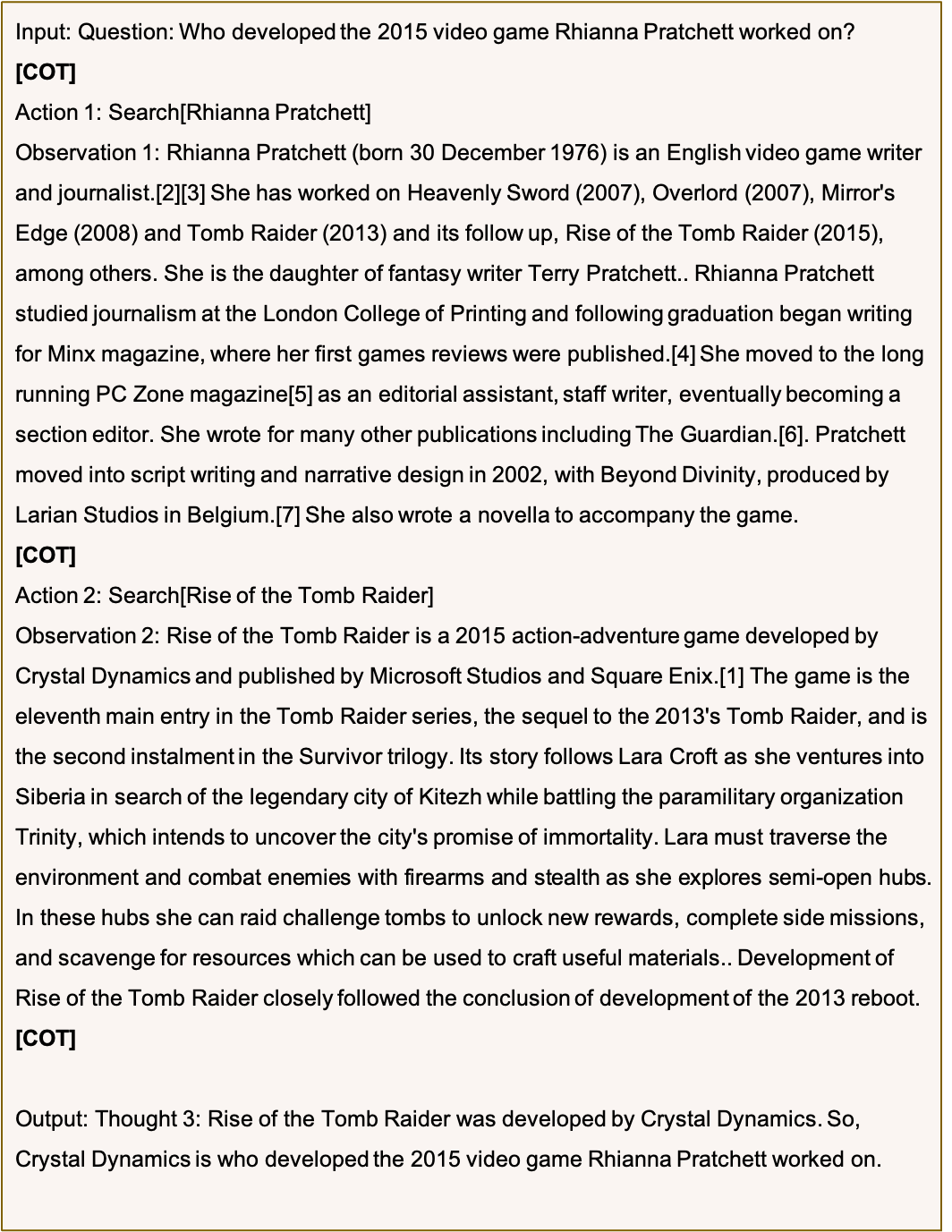}
\end{figure}

\subsection{Implementation Details}
\label{appendix:implementation_details}
All models were trained using DeepSpeed's ZeRO-3 \cite{rasley2020deepspeed} with BF16 precision on 32 NVIDIA A800 80GB GPUs. Each GPU device had a train batch size of 1, and training was conducted for up to 20 epochs. We performed gradient accumulation every 8 steps and saved checkpoints every 50 steps. The best checkpoint was selected based on a designated development set for each dataset. This checkpoint was then evaluated on the test set, with the final test set accuracy reported. During inference, we set the temperature to 0.01 and top\_p to 1.
To evaluate the math problem-solving accuracy, we utilized GPT-4 as the answer extractor and verifier. The accuracy of GPT-4's judgments exceeded 95\%. For the ScienceQA tasks, the output pattern was highly observable, and we recognized the final answer using regular expressions. For agent invocation tasks, we employed a similar approach, utilizing regular expressions to recognize patterns such as "Action \d+: Finish | search | lookup."
By leveraging state-of-the-art techniques and carefully curated evaluation procedures, we aimed to conduct a comprehensive and rigorous assessment of our proposed method's performance across various task domains.